\documentclass[sigconf]{acmart}
\renewcommand\footnotetextcopyrightpermission[1]{} 
\usepackage{times}
\usepackage{helvet}
\usepackage{courier}
\usepackage{graphicx}
\usepackage{subfigure}
\usepackage{verbatim}
\usepackage{url}
\usepackage{amsmath}
\usepackage{balance}
\usepackage{tabularx}
\usepackage{makecell}

\AtBeginDocument{
  \providecommand\BibTeX{{
    \normalfont B\kern-0.5em{\scshape i\kern-0.25em b}\kern-0.8em\TeX}}}

 \setcopyright{acmcopyright}
 \copyrightyear{2018}
 \acmYear{2018}
 \acmDOI{10.1145/1122445.1122456}

\acmConference[MM '20]{Proceedings of the 28th ACM International Conference on Multimedia (MM '20)}{October 12--16, 2020}{Seattle, United States}
\acmBooktitle{the 28th ACM International Conference on Multimedia (MM '20),
  October 12--16, 2020, Seattle, United States}
\acmPrice{15.00}
\acmISBN{978-1-4503-XXXX-X/18/06}

\acmSubmissionID{123-A56-BU3}

\begin{document}

\title{Do You Live a Healthy Life? Analyzing Lifestyle by Visual Life Logging}

\author{Qing Gao, Mingtao Pei, Hongyu Shen}
\affiliation{
  \institution{Beijing Laboratory of Intelligent Information Technology\\Beijing Institute of Technology, Beijing 100081, P.R. China}
 }
\begin{abstract}
A healthy lifestyle is the key to better health and happiness and has a considerable effect on quality of life and disease prevention. Current lifelogging/egocentric datasets are not suitable for lifestyle analysis; consequently,  there is no research on lifestyle analysis in the field of computer vision. In this work, we investigate the problem of lifestyle analysis and build a visual lifelogging dataset for lifestyle analysis (VLDLA). The VLDLA contains images captured by a wearable camera every 3 seconds from 8:00 am to 6:00 pm for seven days. In contrast to current lifelogging/egocentric datasets, our dataset is suitable for lifestyle analysis as images are taken with short intervals to capture activities of short duration; moreover, images are taken continuously from morning to evening to record all the activities performed by a user. Based on our dataset, we  classify the user activities in each frame and use three latent fluents of the user, which change over time and are associated with activities, to measure the healthy degree of the user's lifestyle. The scores for the three latent fluents are computed based on recognized activities, and the healthy degree of the lifestyle for the day is determined based on the scores for the latent fluents. Experimental results show that our method can be used to analyze the healthiness of users' lifestyles.
\end{abstract}



\maketitle

\section{Introduction}
\noindent A healthy lifestyle is the key to better health and happiness. A healthy lifestyle has a considerable effect on the quality of life and disease prevention and can prolong life expectancy~\cite{Li2018Impact}. However, many people are currently living with unhealthy lifestyles, such as working in front of a computer or playing computer games for long periods of time without rest, not drinking water for a long time, and not eating regularly. An unhealthy lifestyle may lead to many noncommunicable chronic diseases, such as hypertension \cite{Bruno2016Association}, diabetes mellitus\cite{yang2011human} and cardiovascular diseases\cite{tol2013health}, and is a major threat to a healthy and happy life.  

To increase the healthiness of a lifestyle, a person's way of life must first be analyzed. Currently, many apps and various software are available to help people record their daily activities and analyze their lifestyle. However, almost all these apps and software require manual input and are difficult to use. 

The rapid development of wearable devices has resulted in visual lifelogging attracting increasing research attention. Visual lifelogging consists of acquiring images that capture the daily experiences of users who wear a camera over a long period of time and automatically analyzing the activities based on the captured egocentric data~\cite{bolanos2017toward}. Visual lifelogging has a wide range of applications, such as health monitoring~\cite{amin2016curating}, personal data archiving~\cite{gemmell2002mylifebits}, and stimulation for memory rehabilitation~\cite{browne2011sensecam}.  Many visual lifelogging methods have been proposed\cite{yan2015egocentric,wang2016characterizing,doherty2011passively,cartas2017batch}; the current visual lifelogging research focuses on daily activity recognition, such as cooking and working. In this paper, we  analyze the lifestyle of a user based on visual lifelogging, which can help the user to establish a healthy lifestyle. To the best of our knowledge, this is the first attempt to analyze lifestyle by means of visual lifelogging.

To analyze a user's lifestyle via visual lifelogging, the data captured by a wearable device should cover the entire day and be captured at intervals shorter than the duration of any activity of interest. In theory,  continuously capturing video for an entire day is the optimal; however, the currently available wearable devices cannot capture videos for a whole day due to storage and battery limitations.  

\begin{figure}
   \centering
   \includegraphics[width=1.00\linewidth]{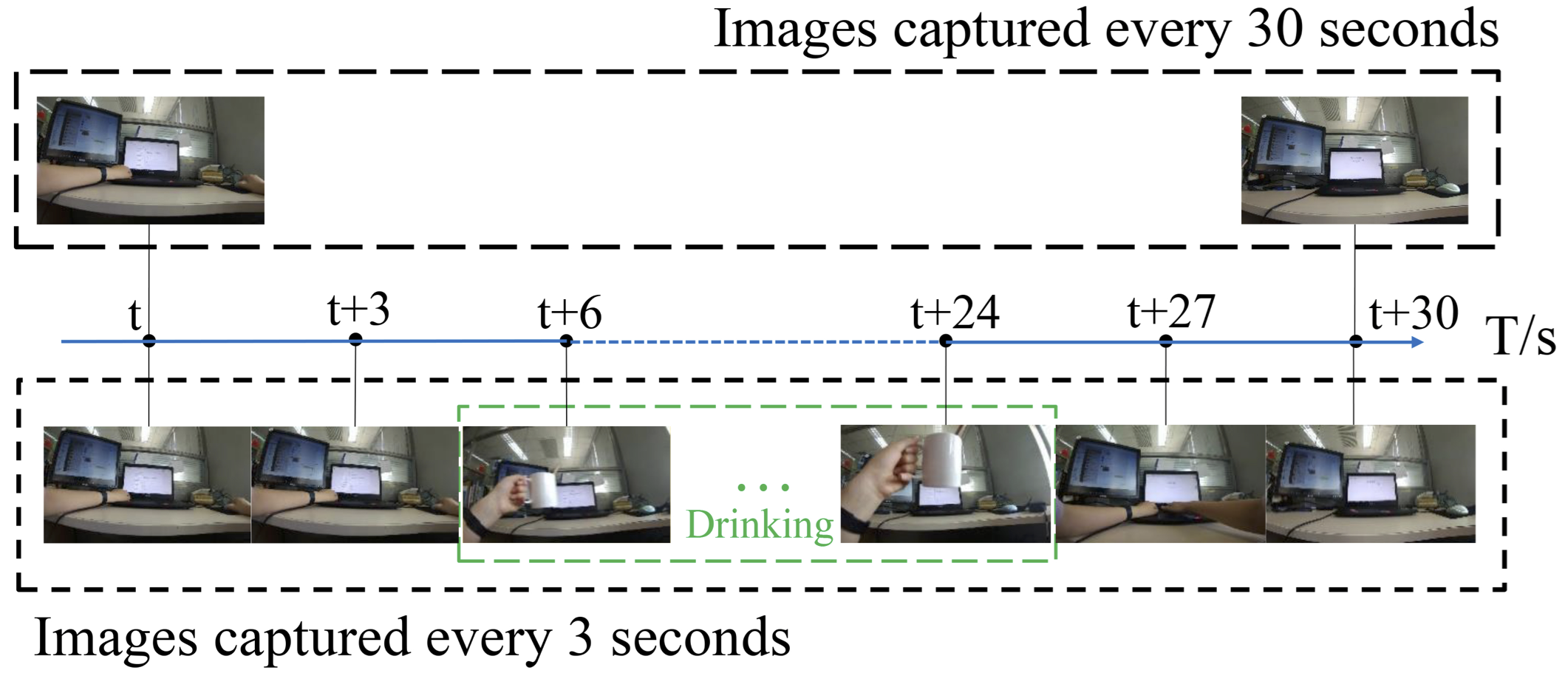}
   \caption{Our dataset contains more details about daily activities, such as drinking, that are missing from datasets with images captured every 30 seconds.} 
   \label{figure1}
\end{figure} 

The currently available lifelogging/egocentric datasets can be classified into two categories: 

1) Datasets with images taken every tens of seconds to record the daily life of a user for a long period of time, such as AIHS~\cite{jojic2010structural}, ImageCLEFlifelog2020~\cite{ImageCLEFlifelog2020} and EDBU~\cite{bolanos2015ego}. As the time interval between images is fairly large, some details of daily activities are missing and some activities with short durations cannot be captured.  For example, Figure \ref{figure1} shows the images captured every 30 seconds and images captured every 3 seconds. If the images are captured every 30 seconds, for example, the two images on the top row of Figure \ref{figure1}, an observer would think the user is using the computer the entire time. However, with images captured every 3 seconds, for example, the images on the bottom row of Figure \ref{figure1},  an observer can see that the user drinks water while using the computer. The drinking water activity is important for lifestyle analysis because, as mentioned previously, failure to drink water for a long time is not consistent with a healthy lifestyle. 

\begin{figure}
   \centering
   \includegraphics[width=1.00\linewidth]{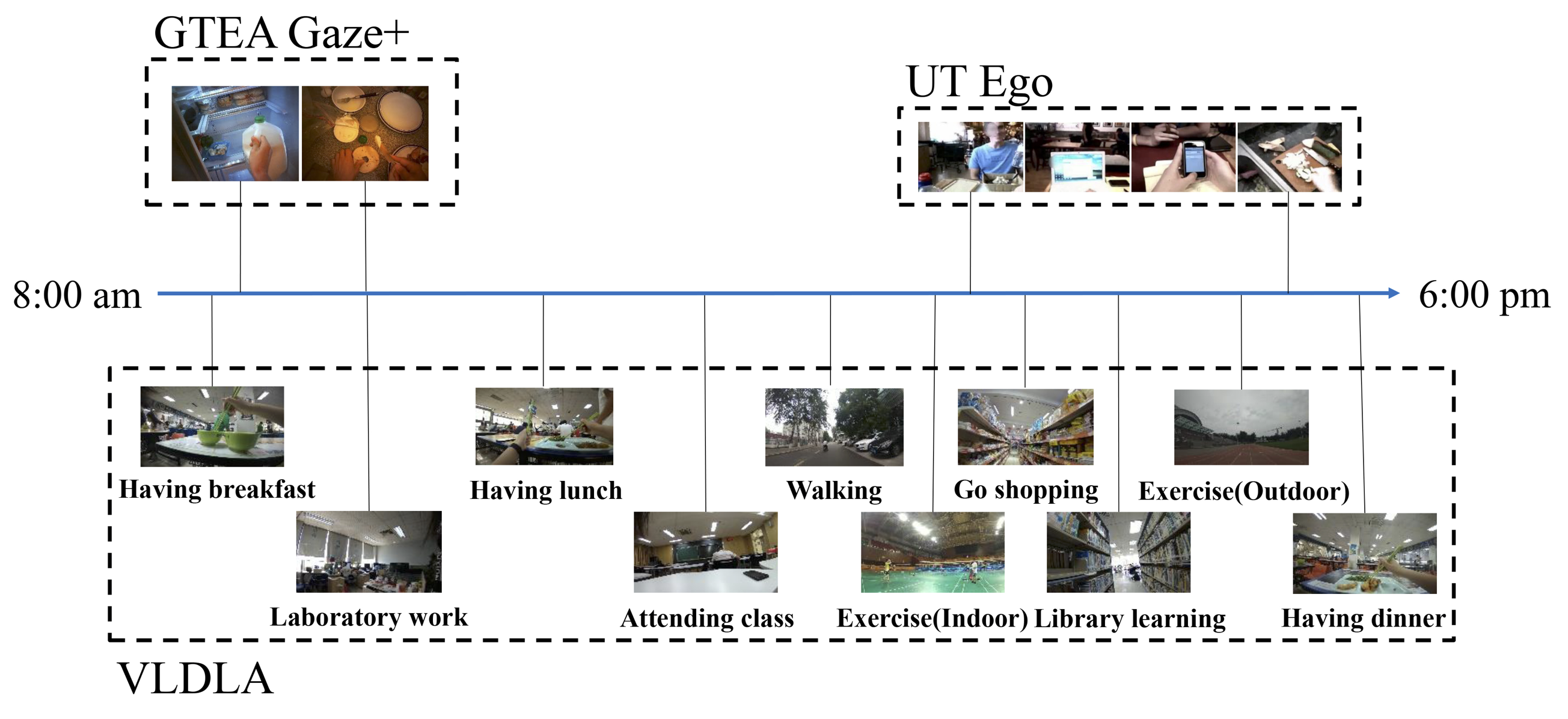}
   \caption{In contrast to datasets with separate videos, our dataset covers the activities of a whole day and is suitable for lifestyle analysis.} 
   \label{figure2}
\end{figure} 

2) Datasets with video clips to record certain activities, with each video clip typically corresponding to a single activity, such as the Extended GTEA Gaze+~\cite{li2018eye},  UT Ego~\cite{lu2013story} and EgoSeg~\cite{poleg2014temporal}. These datasets are not suitable for lifestyle analysis as the video clips do not cover a long period of time, such as a complete day. For example, the top row of Figure \ref{figure2} shows two videos in GTEA Gaze+ and UT Ego. The user is cooking in the first video and working in the second video, but whether the user's lifestyle is healthy cannot be determined by watching these separate videos. On the basis of the second row of Figure \ref{figure2}, one can see that the user takes breaks while working and eats and drinks regularly, which is consistent with a healthy lifestyle.

Current lifelogging datasets are either missing details about activities (because of the large time interval between the captured images) or do not cover a long period of time (video clips of different activities are captured separately) and are thus not suitable for lifestyle analysis. Based on the above observations, we build a visual lifelogging dataset for lifestyle analysis (VLDLA) with images captured with a wearable camera every three seconds from 8:00 am to 6:00 pm for seven days. 

Based on our dataset, we propose three latent fluents to measure the healthiness of a lifestyle. The three latent fluents are \emph{fatigue}, to indicate whether the user rests regularly; \emph{thirst}, to indicate whether the user drinks regularly; and \emph{hunger}, to indicate whether the user eats regularly. The three latent fluents change over time; for example, a working user will become increasingly tired as time goes on. Certain activities can also influence the latent fluents, e.g.,  resting can restore the fatigue fluent and eating can restore the hunger fluent. The overall lifestyle score for a day is computed based on the three latent fluents.

The activities of a user must be recognized first to analyze the three latent fluents. We classify the activities in each frame by combining the scene context, object context and temporal information. The proposed method is based on the following observations.  (1). Specific activities will always occur in certain scenes, and the scene context can provide important information for activity classification. (2). Objects are important for many daily activities, and object context can also help to classify activities. (3). There are temporal correlations between frames of each activity and temporal constraints between different activities, which can be used for activity classification.  Based on the above observations, we propose to extract scene features and object features via ResNet and Mask-RCNN, respectively,  and model the temporal correlations between frames and the  temporal constraints between different activities by means of long short-term memory (LSTM) and conditional random fields (CRFs), respectively. 

The main contributions of this paper can be summarized as follows:

\begin{itemize}
\vspace{0cm}
\item [1] 
This study is the first attempt to analyze a user's lifestyle via visual lifelogging, and we build the VLDLA with images captured every three seconds from 8:00 am to 6:00 pm for seven days. The VLDLA is suitable for lifestyle analysis as it covers all the activities of each day in sufficient detail. 
\item [2]
We analyze the lifestyle based on three latent fluents, namely, fatigue, hunger and thirst, which reflect the user's condition. 
\item [3]
We classify the daily activities of the user based on the scene context, object context, temporal correlations between frames in each activity, and temporal constraints between different activities simultaneously.
\vspace{0cm}
\end{itemize}

\section{Related work}
\subsection{Egocentric Dataset}
Various lifelogging/egocentric datasets have been collected and published in recent years. The EPIC-Kitchens Dataset\cite{damen2018scaling} is a large-scale egocentric video benchmark that contains 55 hours of video recorded by 32 participants in different kitchens. The recordings start immediately before the participants enter the kitchen and stop before the participants leave the kitchen. The Extended GTEA Gaze+ dataset \cite{li2018eye} contains 28 hours of cooking activities from 86 unique sessions of 32 subjects, and annotations of actions (human-object interactions) and hand masks are provided. These two recently released datasets are very large; however, all the activities recorded are performed in kitchens, which is not suitable for lifestyle analysis.

Life-logging EgoceNtric Activities (LENA)~\cite{song2014activity} contains 13 categories of activities with 20 video clips taken by GoogleGlass. The UT Ego Dataset~\cite{lu2013story} has 4 videos, each consisting of 3-5 hours of continuous recording of daily lives with Looxcie. Charades-Ego \cite{sigurdsson2018charades} contains 34.4 hours of first-person videos consisting of 68,536 activity instances, and it is the only large-scale dataset to offer pairs of first- and third-person views. However, these datasets are not suitable for lifestyle analysis because the video clips do not cover activities over a long period of time.

All I Have Seen (AIHS)~\cite{jojic2010structural} contains 45,612 images taken with SenseCam every 20 seconds for 19 days. ImageCLEFlifelog2020~\cite{ImageCLEFlifelog2020} provides a multimodal dataset that consists of approximately 4.5 months  of data from three lifeloggers, including images (1,500-2,500 per day from wearable cameras), semantic content (semantic locations, semantic activities) based on sensor readings on mobile devices, and biometric information. In these datasets, as the time interval between the images is fairly long, some details of the daily activities are missing and some activities with short duration cannot be captured. 

Therefore, current lifelogging/egocentric datasets are not suitable for lifestyle analysis, as they are either missing details of activities or do not cover a long period of time.

\subsection{Egocentric Activity Analysis}
Many methods have been proposed for egocentric activity analysis. Fathi et al. \cite{fathi2011understanding} propose a probabilistic model to map activities into a set of actions and to model each action as a spatiotemporal relationship between the hands and the involved objects. Ma et al.\cite{Ma2016Going} propose a twin-stream CNN architecture for first-person activity recognition. The two streams use the appearance of the object of interest and the optical flow to recognize the object and action, respectively, and the activity is identified by combining the two streams. Singh et al.\cite{Singh2016First} propose a  four-stream neural network architecture that combines temporal egocentric features, spatial information and optical flow to recognize first-person actions.  

Yan et al. ~\cite{yan2015egocentric} propose to recognize similar actions in similar environments by means of a multitask clustering framework based on the observation that the tasks of recognizing everyday activities of multiple individuals are related. Wang et al. ~\cite{wang2016characterizing} use semantic concepts in images to autogenerate summaries of daily activities. They characterize the everyday activities and behavior of subjects by applying a hidden conditional random field (HCRF) algorithm to an enhanced representation of semantic concepts appearing in visual lifelogs.

Castro et al. \cite{castro2015predicting} combine the classification probabilities of a CNN with time and global features, namely, a color histogram, through a random decision forest to classify images into 19 different activity categories. Cartas et al. \cite{cartas2017recognizing} use the outputs of different layers of a CNN as contextual information instead of using color and time information, which are strongly tied to a single user context. Oliveira et al. \cite{oliveira2017leveraging} use a gradient boosting machine approach to retrieve activities based on their estimated relations with objects in the scene.  Poleg et al. ~\cite{poleg2014temporal} address the motion cues for video segmentation and segment egocentric videos into a hierarchy of motion classes using novel cumulative displacement curves.  

Cartas et al. \cite{cartas2018batch} propose an fine-tuned approach that takes into account the temporal coherence of egocentric photo streams. In their method, LSTM units are added on top of a CNN for each frame that is trained by processing the photo streams using batches of fixed size. Swathikiran Sudhakaran et al. \cite{Sudhakaran_2019_CVPR} propose an end-to-end two-stream long short-term attention (LSTA) network that extends LSTM with recurrent attention and output pooling for egocentric activity recognition.  Evangelos Kazakos et al. \cite{Kazakos_2019_ICCV} propose an end-to-end trainable audio-visual temporal binding network(TBN) for egocentric action recognition.

Most current egocentric activity recognition methods use a CNN to extract frame features and LSTM to capture the temporal correlations between frames; however, temporal constraints between activities are not considered.  

\section{Dataset}
\begin{figure}
   \centering
   \includegraphics[width=0.85\linewidth]{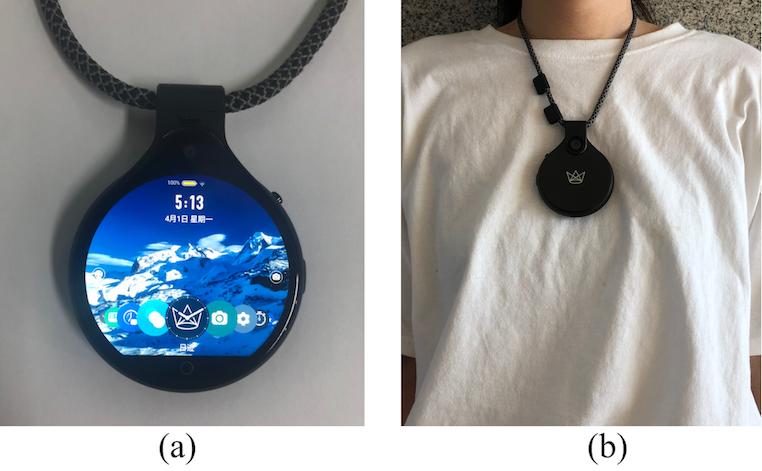}
   \caption{The shooting tool. (a) A FrontRow. (b) A user wearing a FrontRow.} 
   \label{figure3}
\end{figure}

We build the VLDLA by means of a FrontRow wearable camera, which is a neck camera launched by Ubiquiti in 2017. FrontRow is a portable wearable lifelogging device that can be worn on the chest for 147.5 $^{\circ}$ wide-angle shooting. We select FrontRow because compared with wearable devices such as GoPro, Google Glass, and SenseCam, FrontRow can record daily lives more clearly, comprehensively and naturally. According to previous research~\cite{bolanos2017toward}, devices worn on the chest are considered to be the best choice for recording lifelogging data, in contrast to devices placed on other body parts such as the head and eyes, as devices placed on the chest can capture more data and have the least impact on daily lives. Figure~\ref{figure3} shows the FrontRow camera we used. 

\begin{figure}[t]
   \centering
   \includegraphics[width=0.85\linewidth]{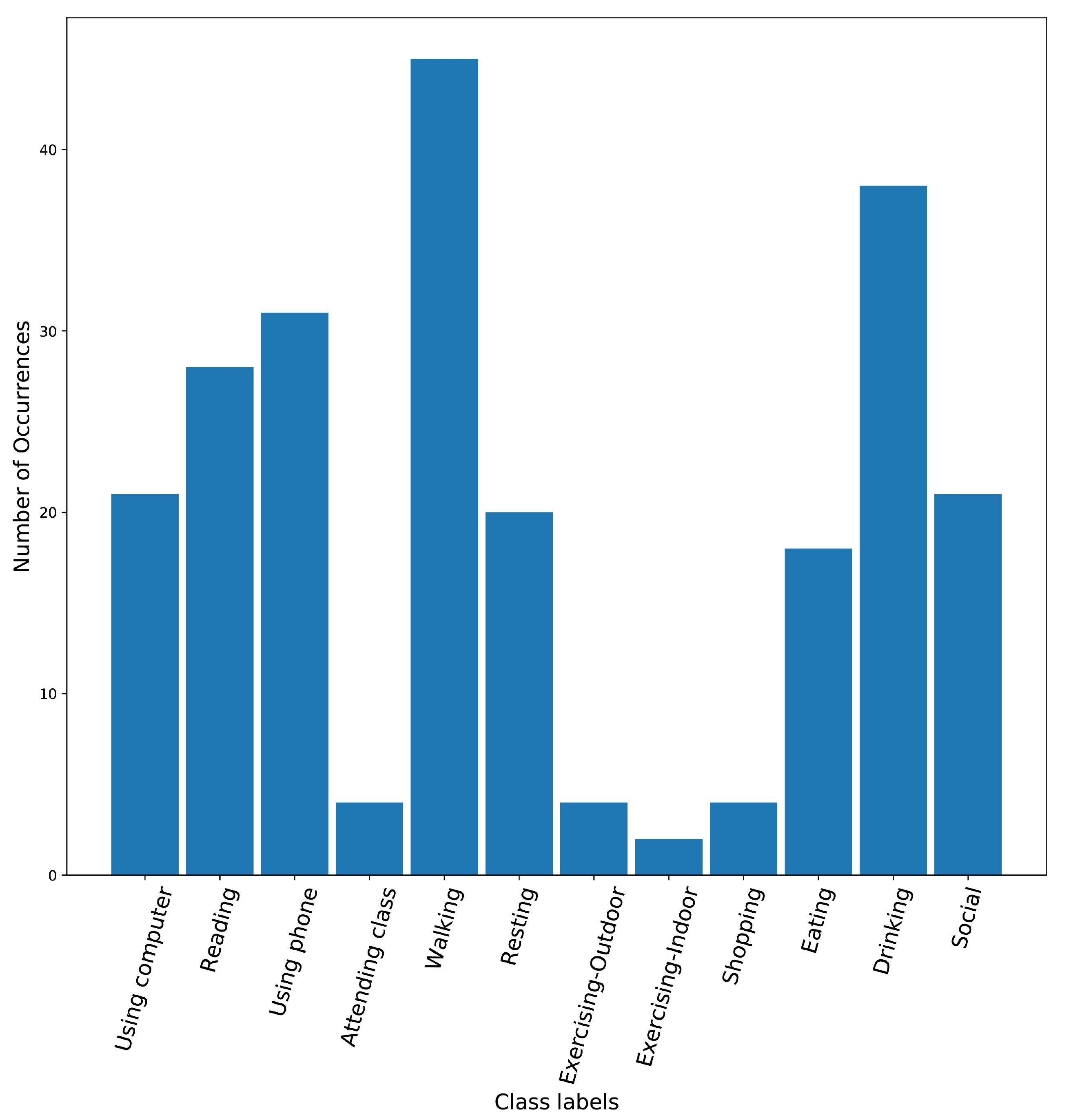}
   \caption{Number of occurrences of each activity in VLDLA. 
   } \label{activitytree}
\end{figure}

\begin{figure*}[t]
   \centering
   \includegraphics[width=0.95\linewidth]{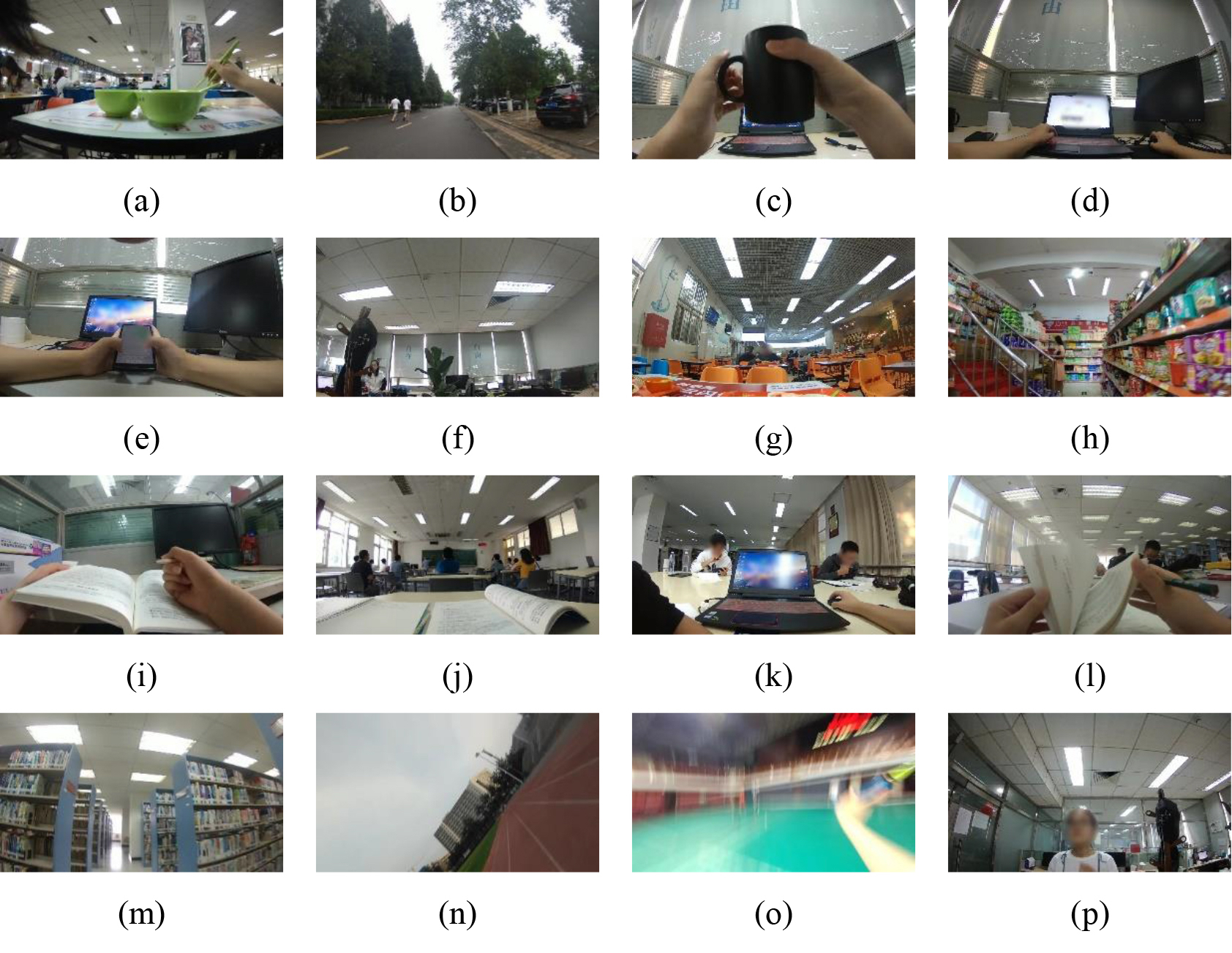}
   \caption{Sample frames from our VLDLA dataset. The activities in the frames are (a),(g): eating, (b):walking, (c): drinking, (d),(k): using computer, (e): using phone, (f),(m): resting, (h): shopping,  (i),(l): reading, (j): attending class, (n): exercising-outdoor, (o): exercising-indoor,(p): social. Note that the faces and screens in the frames are blurred for privacy protection.} 
   \label{figure4}
\end{figure*} 

\begin{figure*}[t]
   \centering
   \includegraphics[width=1.0\linewidth]{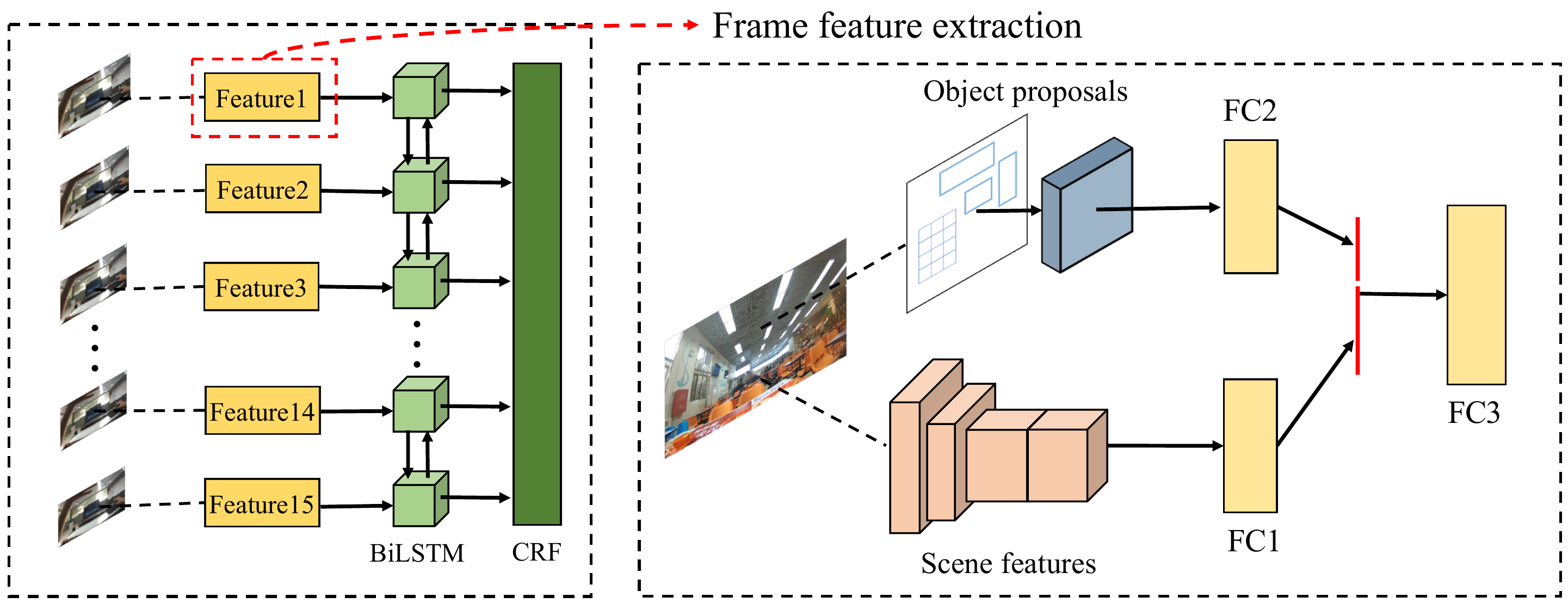}
   \caption{Activity recognition based on scene features, object features and temporal constraints. Scene features from ResNet50 and object proposals from Mask R-CNN are fused as frame feature, and is fed into a BiLSTM-CRF for activity recognition.
    } \label{figure11}
\end{figure*}

\begin{figure}[t]
   \centering
   \includegraphics[width=0.8\linewidth]{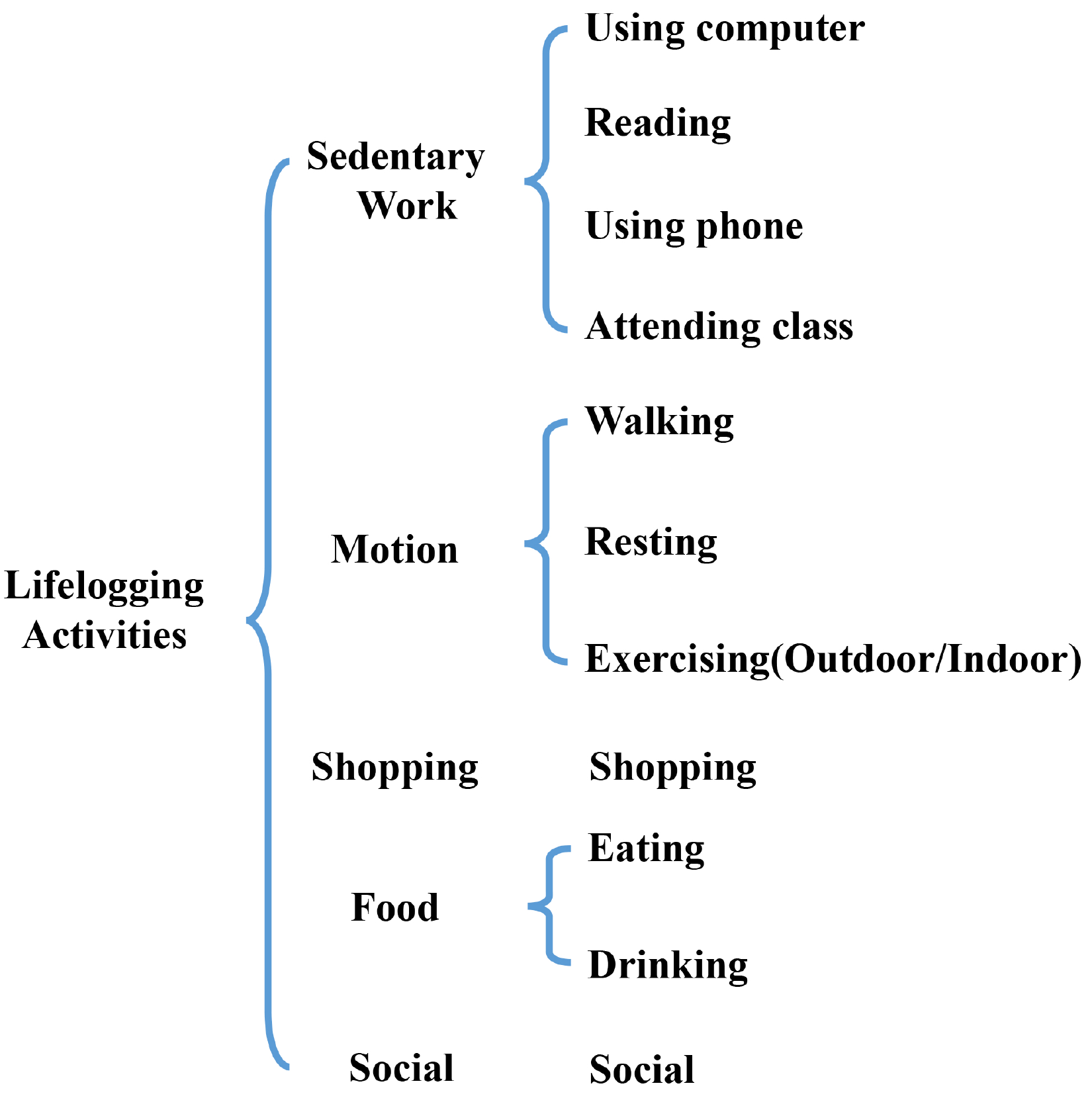}
   \caption{Hierarchical grouping of lifelogging activities in VLDLA. 
   } \label{figure7}
\end{figure}

The VLDLA contains 84,000 images with a resolution of 1920$\times$1080 captured every three seconds from 8:00 am to 6:00 pm for seven days. There are twelve distinct activities: using computer, reading, using phone, attending class, walking, resting, exercising-outdoor, exercising-indoor, shopping, eating, drinking and social. Figure~\ref{activitytree} shows the number of occurrences of each activity in the VLDLA.

For lifestyle analysis, some activities, such as using computer, using phone and reading, are not substantially different as they are all sedentary activities. Therefore, we categorize the twelve activities into 5 groups, as shown in Figure~\ref{figure7}. Intuitively, for a healthy lifestyle, sedentary activities should not last excessively long periods of time, and food-related activities such as eating and drinking should occur regularly throughout the day. As shown in Figure~\ref{figure5}, sedentary activities, food and motion are more well distributed for a healthy lifestyle then an unhealthy lifestyle.

Privacy is a major concern for lifelogging data. We use the face detector in \cite{Zhang2016Joint} and screen detector in \cite{korayem2016enhancing} to detect the faces and screens in each frame and smooth the detected faces and screens with Gaussian filters using a sufficiently large variance\cite{Ribaric2016De} to protect the privacy of the people captured in our dataset. As shown in Figure~\ref{figure4},  the faces and screens in the frames are detected and blurred for privacy protection.

Each frame in the VLDLA is annotated as one of the twelve activities by the user who records the data, as no one can have a better understanding of what the user is doing. Lifestyle is usually defined based on a long period of time. Here, we analyze the lifestyle each day and assign a score from 0 to 1 to indicate whether the lifestyle is healthy: 1 indicates a perfectly healthy lifestyle, and 0 indicates a absolutely unhealthy lifestyle. We ask 10 participants to score the lifestyle for each day by showing them the script generated by the labels of each frame. An example script is: a user uses a computer from 9 am to 11 am, drinks water, uses a computer from 11 am to 12 am, eats for 20 mins, and so on. Then, the average score for each day is computed as the ground truth.

\section{Method}

\subsection{Overview}

Given frames taken every 3 seconds for a day, the activities in each frame are recognized first. Then, scores for the three latent fluents in each frame are computed, and the overall score for the lifestyle of the day is obtained.

\begin{figure*}
   \centering
   \includegraphics[width=0.95\linewidth]{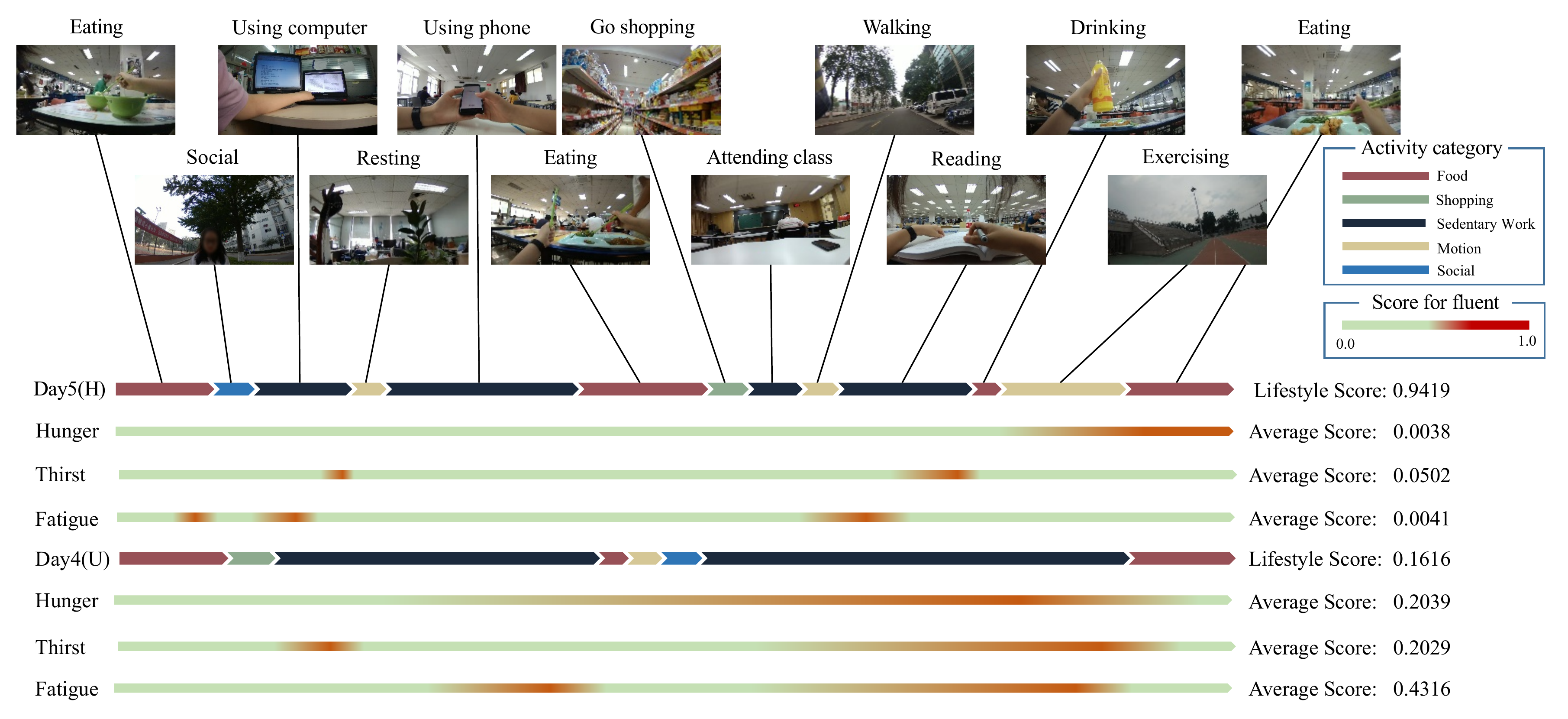}
   \caption{Demonstration of the scores for the three latent fluents and lifestyle. For Day 5, the user has a healthy lifestyle: sedentary work, food and motion are well-distributed, the scores for hunger, thirst and fatigue are low and the score for lifestyle is high. For Day 4, the user participates in excessive sedentary work, and the score for lifestyle is low.} 
   \label{figure5}
\end{figure*}

\subsection{Activity Recognition}
To compute the scores for the latent fluents, the user activities must be recognized first, as many activities can change the latent fluents. We recognize the activity in each frame based on the scene features, object features and temporal information. 

Most of these twelve activities have their own specific occurrence scenarios. For example, eating generally occurs in a cafeteria, restaurant, food\_court, etc. Moreover, walking often occurs outdoors, and eating, laboratory work, attending class and shopping usually occur indoors. We use the ResNet50~\cite{he2016deep} trained on Places365~\cite{zhou2018places} to extract scene features for each frame.

Activities are also generally associated with representative objects, such as saucers and bowls for eating. We use the Mask R-CNN~\cite{he2017mask} trained on COCO to extract object proposals in each frame as object features. 

As shown in Figure~\ref{figure11}, we replace the last original fully connected layer of ResNet 50 with a new fully connected layer (FC1) with a dimension of 500 and treat the 500-dimensional output of the FC1 layer as the scene feature.  We add a fully connected layer (FC2) with a dimension of 400 after the object proposals from the RoiAlign layer of Mask R-CNN and treat the 400-dimensional output of the FC2 layer as the object feature. The scene feature and object feature are concatenated and passed to a fully connected layer (FC3) with a dimension of 500, and the 500-dimensional output of the FC3 layer is used as the combined features.

The combined features are fed into a BiLSTM-CRF\cite{lample2016neural} for activity recognition. The BiLSTM can learn the temporal correlations between the combined features, and the linear-chain CRF can impose temporal constraints between activities, for example, resting usually follows sedentary work and a person will not eat more than once within a short period of time. For the BiLSTM, we use the same implementation as \cite{lample2016neural}

\begin{equation}
\begin{aligned}
i_{t}=\sigma (W_{xi}x_{t}+W_{hi}h_{t-1}+W_{ci}c_{t-1}+b_{i})\\
\end{aligned}
\end{equation} 
\begin{equation}
\begin{aligned}
c_{t}=(1-i_{t})\odot c_{t-1}+i_{t}\odot tanh(W_{xc}x_{t}+W_{hc}h_{t-1}+b_{c})
\end{aligned}
\end{equation} 
\begin{equation}
\begin{aligned}
o_{t}=\sigma (W_{xo}x_{t}+W_{ho}h_{t-1}+W_{co}c_{t}+b_{o})
\end{aligned}
\end{equation} 
\begin{equation}
\begin{aligned}
h_{t}=o_{t}\odot tanh(c_{t})
\end{aligned}
\end{equation} 
where $\sigma$ is the sigmoid function and $\odot$ is the dot product. $h_{t}$ is the output of the BiLSTM and the input of the CRF. We use batches of $n$(15) consecutive frames as the input of the BiLSTM, and the classification result from the CRF is $y=(y_{1}, y_{2}, y_{3}, ... , y_{n})$. The score of $y$ can be computed as
\begin{equation}
\begin{aligned}
s(X,y)=\sum_{i=0}^{n}A_{y_{i},y_{i+1}}+\sum_{i=1}^{n}P_{i,y_{i}}
\end{aligned}
\end{equation} 
where $P$  is the $n*k$ matrix of scores output by the BiLSTM, $k$ is the number of activity classes, and $P_{i,j}$ corresponds to the score of the $j$th label of the $i$th frame in a video. $A_{y_{i},y_{j}}$ represents the transition score between label $y_{i}$ and label $y_{j}$. For more details, please refer to\cite{lample2016neural}.

\subsection{Lifestyle Analysis}
We define three latent fluents to compute the lifestyle score: hunger, thirst, and fatigue. We analyze the lifestyle according to commonly accepted assumptions about a healthy lifestyle, that is, one should take a break after sedentary work and one should eat and drink regularly. We compute a score for each latent fluent in each frame, and the overall lifestyle score is computed as

\begin{equation}
s_{j}^{lifestyle}= 1- \frac{1}{3N_{j}}\sum\limits_{i = 1}^{N_{j}} {(s_{i}^{hunger}+s_{i}^{thirst}+s_{i}^{fatigue})}
\end{equation} 

where $N_{j}$ is the total number of frames for the $jth$ day and $3$ is used to normalize the lifestyle score to $(0,1)$. We use 1 minus the average score of the three latent fluents to make the lifestyle score accord with the convention that 1 corresponds to a healthy lifestyle and 0 corresponds to an unhealthy lifestyle. $s_{i}^{hunger}$, $s_{i}^{thirst}$ and $s_{i}^{fatigue}$ are the scores for the three latent fluents in frame $i$. 

$s_{i}^{hunger}$ is computed as

\begin{equation}
s_{i}^{hunger}=\left\{ \begin{aligned}
 & \qquad \qquad \quad 0 \qquad \qquad \qquad if \quad c_{i} = eating \\
 & \frac{1}{1+e^{-(\frac{i-i_{eating}}{1200}-\alpha_{hunger})}} \quad else\\
\end{aligned} \right.
\end{equation}

where $c_{i}$ is the activity occurring in frame $i$, $i_{eating}$ is the frame index of the last eating activity, $1200$ is the number of frames captured in an hour, and $\alpha_{hunger}$ is a hyperparameter that is set to 5 since a person will typically become hungry approximately 5 hours after eating. $s_{i}^{hunger}$ will be $0.5$ 5 hours after eating and will increase further as time passes. 

$s_{i}^{thirst}$ and $s_{i}^{fatigue}$ are computed similarly to $s_{i}^{hunger}$, and  $\alpha_{thirst}$ and $\alpha_{fatigue}$ are defined similarly to  $\alpha_{hunger}$ and are set to $2$ and $1$, respectively.

\begin{table}
\small
\centering
\setlength{\tabcolsep}{0.5mm}{
\caption{Overall comparison results of activity recognition on VLDLA}
\label{tableoverall}
\begin{tabular}{c|c|c|c|c}
\hline  
Method  & Accuracy & \makecell[c]{Macro \\Precision} & \makecell[c]{Macro \\Recall} & \makecell[c]{Macro \\F1-score}\\  
\hline   
InceptionV3+RF+LSTM~\cite{cartas2018batch}   &0.5669 & 0.6311 & 0.4869 & 0.4615\\ 
Our method   &0.8557 & 0.7695 & 0.8028 & 0.7615\\   
\hline  
\end{tabular}}
\end{table}

\begin{table*}
\small
\centering
\setlength{\tabcolsep}{0.5mm}{
\caption{Comparison results of the F1-score for each activity category}
\label{tablespecific}
\begin{tabular}{c|c|c|c|c|c|c|c|c|c|c|c|c}
\hline  
Method  & Social & \makecell[c]{Using \\computer} & Reading & \makecell[c]{Using \\phone} & \makecell[c]{Attending \\class} & Walking & \makecell[c]{Resting}  & \makecell[c]{Exercising\\(outdoor)}  &  \makecell[c]{Exercising\\(indoor) } & Shopping & Eating & Drinking \\  
\hline  
InceptionV3+RF+LSTM~\cite{cartas2018batch}   &0.5356 & 0.7039 & 0.6045 & 0.0587 & 0.7501 & 0.6717 & 0.5240  & 0.5198 & 0.0267 & 0.3810 & 0.6591 & 0.1032 \\  
Our method   &0.4631 & 0.9064 & 0.8300 & 0.4548 & 0.6488 & 0.8954 & 0.9757  & 0.9593 & 0.9628 & 0.6990 & 0.9564 & 0.3858 \\
\hline  
\end{tabular}}
\end{table*}

\section{Experiment}

\subsection{Activity Recognition}
We split the VLDLA dataset into a training set and testing set. The data from the first, second, third, and sixth days are chosen as the training data, and the remaining data are used for testing. 

\begin{figure}[t]
   \centering
   \includegraphics[width=0.9\linewidth]{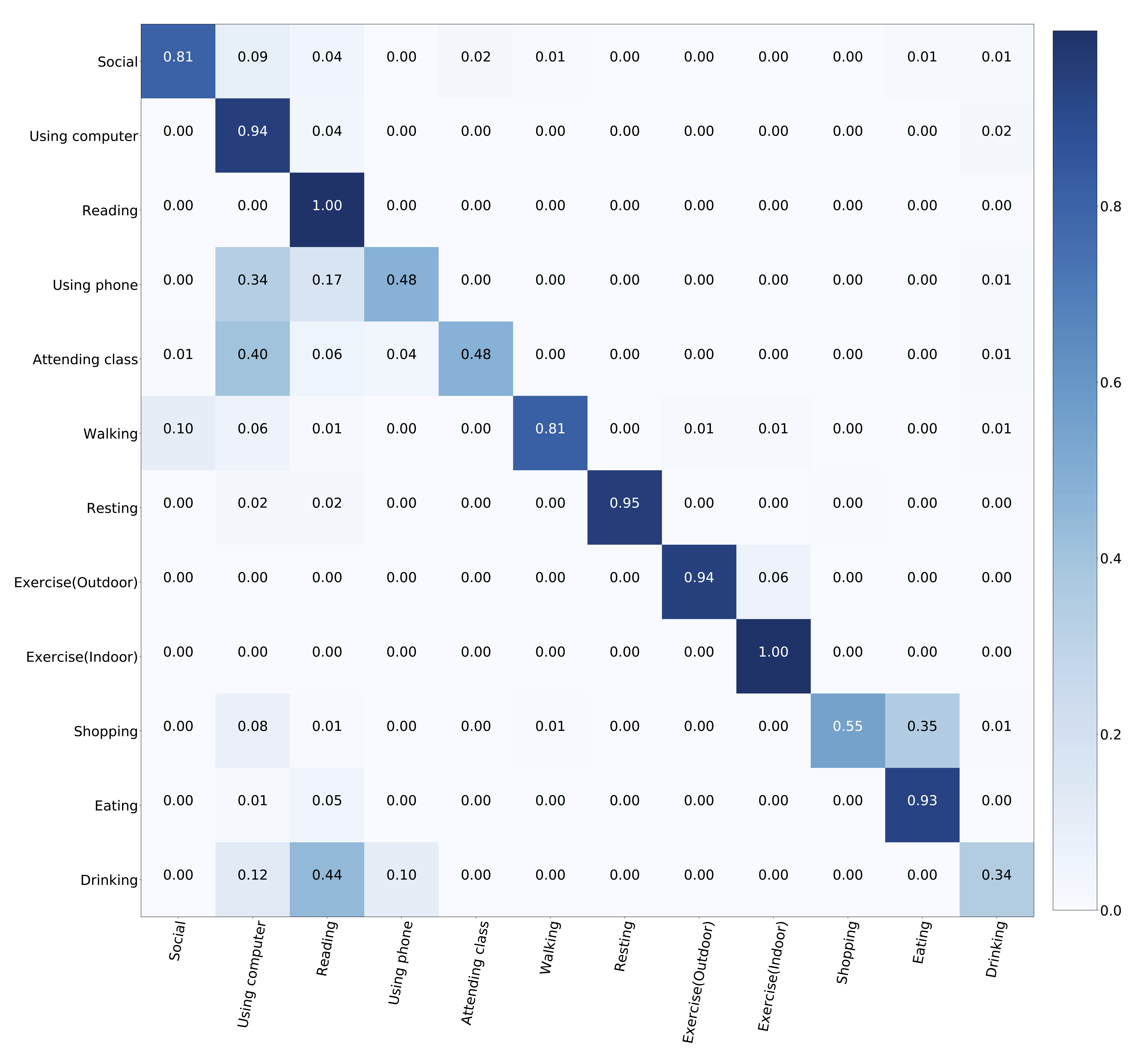}
   \caption{The confusion matrix for the twelve activities.} 
   \label{figure8}
\end{figure}

Figure~\ref{figure8} shows the computed confusion matrix of the twelve activities in our dataset: the combination of scene features, object features and temporal information can be used to effectively classify most of the twelve activities. However, the drinking and using phone activities are not classified very well. One potential explanation is that for these activities, the objects involved are usually difficult to detect because they are usually small and often move out of view.  

\begin{figure}[t]
   \centering
   \includegraphics[width=0.9\linewidth]{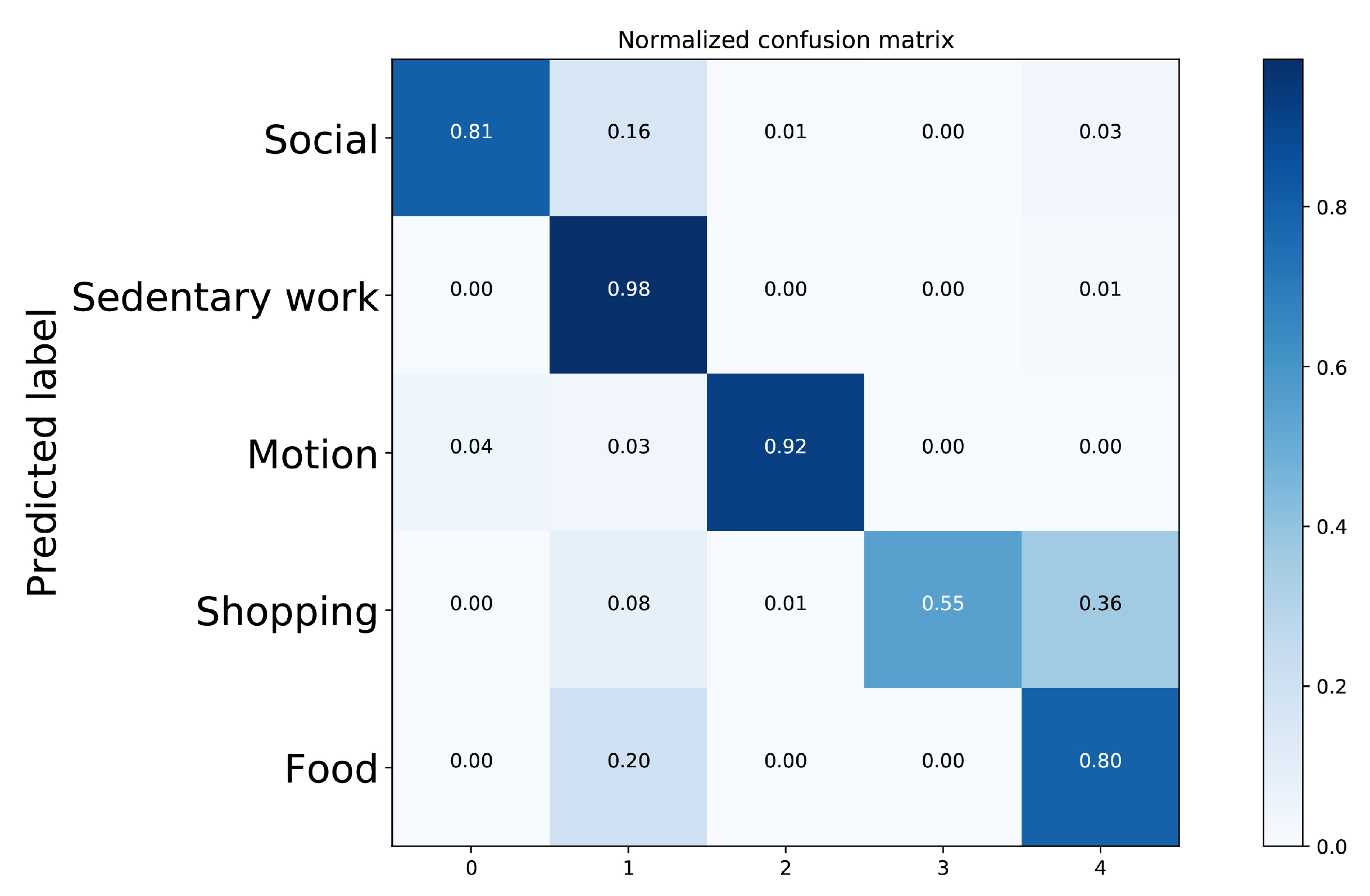}
   \caption{The confusion matrix for the five activity groups.} 
   \label{figure9}
\end{figure}

We also compute the confusion matrix of the five activity groups, as shown in Figure~\ref{figure9}: the five activity groups can be classified clearly. 

We compare our method with InceptionV3+RF+LSTM in~\cite{cartas2018batch}, which achieves the best performance among the many methods compared in ~\cite{cartas2018batch}. The comparison of the accuracy, macro precision, macro recall and macro F1-score is shown in Table~\ref{tableoverall}, and the comparison results of the macro F1-score for each activity category are shown in Table~\ref{tablespecific}. Table~\ref{tableoverall} and Table~\ref{tablespecific} illustrate that our method achieves better performance on most categories because the BiLSTM-CRF can simultaneously impose temporal correlations between the frames in each activity and temporal constraints between different activities.

\subsection{Lifestyle Analysis}

As we mentioned before, we ask 10 participants, including 4 females and 6 males, from a local university, whose ages range from 18 to 25 years, to score the lifestyle for each day based on the script generated by the labels of each frame. An example script is as follows: a user uses a computer from 9 am to 11 am, drinks water, uses a computer from 11 am to 12 am, eats for 20 mins, and so on. Then, the average score for each day is computed as the ground truth.

\begin{figure}
   \centering
   \includegraphics[width=0.9\linewidth]{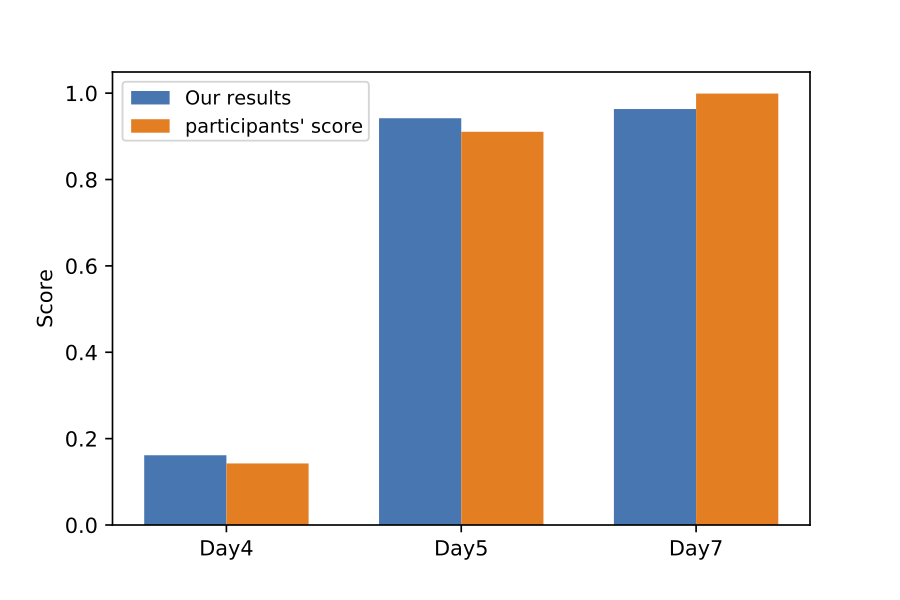}
   \caption{Experimental results for the lifestyle analysis.} 
   \label{figure10}
\end{figure} 

Based on the activity recognition results, we compute the scores for the three latent fluents and for the overall lifestyle according to Equations (6) and (7). Figure~\ref{figure5} shows the demonstration of the scores for the three latent fluents and lifestyle. For Day 5, the user has a healthy lifestyle: sedentary work, food and motion are well-distributed, the scores for hunger, thirst and fatigue are low and the score for lifestyle is high. For Day 4, the user participates in excessive sedentary work, and the score for lifestyle is low. We compare the computed scores with the participants' scores as shown in Figure~\ref{figure10}, where the lifestyle of Day 4 is unhealthy, and the lifestyle of Day 5 and Day 7 is healthy. The computed score for lifestyle accords with the participants' score, which confirms the rationality and effectiveness of our method. 

\section{Conclusion and Discussion}
In this work, we build a new visual lifelogging dataset for lifestyle analysis (VLDLA) that contains images taken every three seconds for seven days. The VLDLA covers a long period of time with images captures at short time intervals and is suitable for lifestyle analysis. Based on the VLDLA, we propose a method for lifestyle analysis based on three latent fluents and the recognition of daily activities.

Currently, our dataset contains only images taken between 8:00 am and 6:00 pm: the data between 6:00 pm and 8:00 am is not captured for privacy concern, as most of activities that occur during this period of time occur in a dormitory.  Our future work includes collecting more data about eating and food to analyze eating habits and collecting multimodal data such as heart rate and blood pressure to analyze the lifestyle more effectively.

\balance
\bibliographystyle{ACM-Reference-Format}
\bibliography{sample-base}

\end{document}